\documentclass[runningheads,a4paper]{llncs}

\usepackage{amssymb, amsmath}
\usepackage{graphicx}
\usepackage{subcaption}
\usepackage{hyperref}
\usepackage{cleveref}
\usepackage{threeparttable}

\usepackage{tikz}

\usepackage{etoolbox}

\apptocmd{\TPTnoteSettings}{\footnotesize}{}{}

\usepackage{url}
\urldef{\mailsa}\path|{lee.clement@mail., v.peretroukhin@mail., j.kelly@utias.}utoronto.ca|    
\newcommand{\keywords}[1]{\par\addvspace\baselineskip
\noindent\keywordname\enspace\ignorespaces#1}

\graphicspath{{figs/}}

\newcommand{\mbf}[1]{\mathbf{#1}}

\newcommand{\mbfhat}[1]{\hat{\mbf{#1}}}

\DeclareMathAlphabet{\mbfh}{OML}{cmm}{b}{it}

\newcommand{\cframe}[1]{\ensuremath \underrightarrow{\mathcal{F}}_{#1}}

\newcommand{\bbm}{\begin{bmatrix}}
\newcommand{\ebm}{\end{bmatrix}}

\newcommand{\set}[1]{\left\{#1\right\}}

\newcommand{\SE}[1]{SE(#1)}

\newcommand\T{\rule{0pt}{2.6ex}}        
\newcommand\B{\rule[-1.2ex]{0pt}{0pt}}  
\begin{document}

\mainmatter  

\newcommand{\titlestring}{Improving the Accuracy of Stereo Visual Odometry Using Visual Illumination Estimation}
\newcommand{\shorttitlestring}{Stereo Visual Odometry Using Visual Illumination Estimation}
\title{\titlestring}

\titlerunning{\shorttitlestring}

%
%
\author{Lee Clement \and Valentin Peretroukhin \and Jonathan Kelly}
\authorrunning{\shorttitlestring}

\institute{Institute for Aerospace Studies,\\
University of Toronto, Toronto, Canada\\
\mailsa}

%
%

\toctitle{\titlestring}
\tocauthor{L Clement, V Peretroukhin, and J Kelly}
\maketitle

\begin{abstract}
In the absence of reliable and accurate GPS, visual odometry (VO) has emerged as an effective means of estimating the egomotion of robotic vehicles.
Like any dead-reckoning technique, VO suffers from unbounded accumulation of drift error over time, but this accumulation can be limited by incorporating absolute orientation information from, for example, a sun sensor.
In this paper, we leverage recent work on visual outdoor illumination estimation to show that estimation error in a stereo VO pipeline can be reduced by inferring the sun position from the same image stream used to compute VO, thereby gaining the benefits of sun sensing without requiring a dedicated sun sensor or the sun to be visible to the camera.
We compare sun estimation methods based on hand-crafted visual cues and Convolutional Neural Networks (CNNs) and demonstrate our approach on a combined 7.8 km of urban driving from the popular KITTI dataset, achieving up to a 43\% reduction in translational average root mean squared error (ARMSE) and a 59\% reduction in final translational drift error compared to pure VO alone.
\end{abstract}

\section{Motivation, Problem Statement, and Related Work}
In the absence of reliable and accurate GPS, visual odometry (VO) has emerged as an effective means of estimating the egomotion of robotic vehicles as they navigate through their environment.
While VO is generally less prone to drift than other dead-reckoning techniques such as wheel odometry, any dead-reckoning algorithm will inevitably accumulate drift over time due to the compounding of small estimation errors.
Indeed, VO suffers from superlinear growth of drift error with distance travelled, mainly due to error in the orientation estimates~\cite{Olson2003-ax}.
Fortunately, the addition of absolute orientation information from, for example, a sun sensor can restrict this growth to be linear~\cite{Olson2003-ax}.

The sun is an appealing source of absolute orientation information since it is readily detectable and its apparent motion through the sky is well characterized in ephemeris tables.
The benefits of deriving orientation information from a sun sensor have been successfully demonstrated in planetary analogue environments~\cite{Furgale2011-zu,Lambert2012-um} as well as on board the Mars Exploration Rovers (MERs)~\cite{Eisenman2002-cg,Maimone2007-tc}.
In particular, Lambert et al.~\cite{Lambert2012-um} showed that by incorporating sun sensor and inclinometer data directly in a stereo VO pipeline, the accumulated drift error can be greatly reduced compared to pure VO alone.

In this work, we seek to answer the question of whether similar reductions in stereo VO drift can be obtained solely from the image stream already being used to compute VO.
The main idea here is that by reasoning over more than just the geometric information available from a standard RGB camera, we can improve existing VO techniques without needing to rely on a dedicated sun sensor or specially oriented camera.
Recently, Lalonde et al.~\cite{Lalonde2011-jw} demonstrated that the likely direction of the sun can be estimated from a single RGB image using a combination of weak visual cues such as shadows and a model of the sky~\cite{Perez1993-cy}.
We improve the accuracy and reliability of this technique by incorporating information from the VO estimate itself, and combine it with a modified version of the sun-sensor-augmented stereo VO pipeline developed by Lambert et al.~\cite{Lambert2012-um} to show that VO drift error can be reduced in this way.
We also investigate the use of a recent machine learning approach to sun direction estimation, which makes use of a Convolutional Neural Network (CNN) to predict the azimuth angle of the sun~\cite{Ma2016-zl}.
We present experimental results demonstrating our approach on a combined 7.8 km of urban driving from the popular KITTI dataset~\cite{Geiger2013-ky}, achieving up to a 59\% reduction in final translational drift error and a 43\% reduction in translational average root mean squared error (ARMSE) compared to pure VO.

\section{Technical Approach}
We adopt a sliding window stereo VO technique that has been used in a number of successful mobile robotics applications~\cite{Cheng2006-nl,Furgale2010-to,Geiger2011-xe,Kelly2008-mh}.
While this technique is not the absolute state of the art,\footnote{Results from several state-of-the-art VO systems on the KITTI odometry benchmark can be found at \url{http://www.cvlibs.net/datasets/kitti/eval_odometry.php}.} it serves as an easily implementable baseline system against which to evaluate our use of visual illumination estimation in the VO pipeline.
We stress that our main idea is not tied to any specific VO technique and could be used in any VO system where RGB images are available.

Our goal is to estimate a window of $\SE{3}$ poses $\set{\mbf{T}_{k+1,b}, \dots, \mbf{T}_{k+N,b}}$ expressed in a base coordinate frame $\cframe{b}$, which we choose to be the first pose in each window.
Our VO pipeline tracks keypoints across pairs of stereo images and computes an initial guess for each pose in the window using frame-to-frame point cloud alignment, which it then refines using a local bundle adjustment over the window.
Finally, the estimated camera trajectory can be transformed into a desired world coordinate frame $\cframe{w}$ given the transformation $\mbf{T}_{b,w}$, which can be obtained from the bundle adjustment solution of the previous window.
As we discuss in \Cref{sec:orientation}, we select the initial pose $\mbf{T}_{1,w}$ to be the first GPS ground truth pose such that $\cframe{w}$ is a local East-North-Up (ENU) coordinate system.

\subsection{Observation Model}
We assume that our stereo images have been de-warped and rectified in a pre-processing step, and model the stereo camera as a pair of perfect pinhole cameras with focal lengths $f_u, f_v$ and principal points $\left(c_u,c_v\right)$, separated by a fixed and known baseline $\ell$.
If we take $\mbf{p}_b^j$ to be the homogeneous 3D coordinates of keypoint $j$, expressed in our chosen base frame $\cframe{b}$, we can transform the keypoint into the camera frame at pose $k$ to obtain $\mbf{p}_k^j = \mbf{T}_{k,b}\mbf{p}_b^j = \bbm p_{k,x}^j & p_{k,y}^j & p_{k,z}^j & 1 \ebm^T$. 
Our observation model $\mbf{g}\left(\cdot\right)$ can then be formulated as
\begin{align} \label{eq:cam_model}
    \mbf{y}_{k,j} &= \mbf{g}\left(\mbf{p}_k^j\right) 
                   = \mbf{g}\left(\mbf{T}_{k,b} \mbf{p}_b^j\right)
                    = \bbm u \\ v \T \\ d \T \ebm 
                    = \bbm 
			   		    f_u p_{k,x}^j / p_{k,z}^j + c_u \\
			   		    f_v p_{k,y}^j / p_{k,z}^j + c_v \T \\
			   		    f_u \ell / p_{k,z}^j  \T
			         \ebm,
\end{align}
where $\left(u,v\right)$ are the pixel coordinates in the left image and $d$ is the disparity.

\subsection{Sliding-window Visual Odometry} \label{sec:technialapproach:vo}
We use the open-source \texttt{libviso2} package~\cite{Geiger2011-xe} to detect and track keypoints between stereo image pairs.
Based on these keypoint tracks, a three-point Random Sample Consensus (RANSAC) algorithm~\cite{Fischler1981-ue} generates an initial guess of the interframe motion and rejects outlier keypoint tracks by thresholding their reprojection error.
We compound these pose-to-pose transformation estimates through our chosen window and refine them using a local bundle adjustment, which we solve using the nonlinear least-squares solver Ceres~\cite{ceres-solver}.
The objective function to be minimized can be written as
\begin{equation} \label{eq:cost_function}
    \mathcal{J} = \sum_{k} \sum_{j} \mbf{e}_{\mbf{y}_{k,j}}^T \mbf{R}^{-1}_{\mbf{y}_{k,j}} \mbf{e}_{\mbf{y}_{k,j}},
\end{equation}
where $\mbf{e}_{\mbf{y}_{k,j}} = \mbfhat{y}_{k,j} - \mbf{y}_{k,j}$ is the reprojection error of keypoint $j$ for camera pose $k$, $\mbf{R}_{\mbf{y}_{k,j}}$ is the covariance of the errors, and the outer sum runs over the chosen window of poses.
The predicted measurements are given by $\mbfhat{y}_{k,j} = \mbf{g}\left(\mbfhat{T}_{k,b} \mbfhat{p}^j_{b}\right)$, where $\mbfhat{T}_{k,b}$ and $\mbfhat{p}^j_{b}$ are the estimated poses and keypoint positions in base frame $\cframe{b}$, which we choose to be the first camera frame in the window.

\subsection{Orientation Correction} \label{sec:orientation}
In order to combat drift in the VO estimate produced by accumulated orientation error, we adopt the technique of Lambert et al.~\cite{Lambert2012-um} to incorporate absolute orientation information from the sun directly into the estimation problem.
We assume the initial camera pose and its timestamp are available from GPS and use them to determine the global direction of the sun $\mbf{s}_w$, expressed as a 3D unit vector, from ephemeris data, where we have defined the world frame $\cframe{w}$ to be a local ENU coordinate frame.
For simplicity, we assume that the full trajectory of the camera is sufficiently short so that the sun is effectively static, although it would be straightforward to obtain the global sun direction at each timestep for longer trajectories where the apparent motion of the sun is significant.

By transforming the global sun direction into each camera frame $\cframe{k}$ in the window, we obtain predicted sun directions $\mbfhat{s}_k = \mbfhat{T}_{k,b} \mbf{T}_{b,w} \mbf{s}_w$, where $\mbfhat{T}_{k,b}$ is the current estimate of camera pose $k$ in the base frame, and $\mbf{T}_{b,w}$ is the fixed, previously estimated transformation from the world frame to the base frame. 
We compare the predicted and estimated sun directions to introduce an additional error term into the bundle adjustment cost function (cf. \Cref{eq:cost_function}):
\begin{equation} \label{eq:cost_function_with_sun}
    \mathcal{J} = \sum_{k}  \left(\sum_{j}  \mbf{e}_{\mbf{y}_{k,j}}^T \mbf{R}^{-1}_{\mbf{y}_{k,j}} \mbf{e}_{\mbf{y}_{k,j}} + \mbf{e}_{\mbf{s}_k}^T \mbf{R}^{-1}_{\mbf{s}_k} \mbf{e}_{\mbf{s}_k} \right),
\end{equation}
where $\mbf{e}_{\mbf{s}_k} = \mbfhat{s}_k - \mbf{s}_k$ is the error in the predicted sun direction, and $\mbf{R}_{\mbf{s}_k}$ is the covariance of the errors.
This additional term constrains the orientation of the camera, which helps limit drift in the VO result due to orientation error~\cite{Lambert2012-um}.

In contrast to~\cite{Lambert2012-um}, we operate directly on the 3D unit sun vectors rather than the underlying two angular degrees of freedom.
While we could also use cosine distance as the error term in our cost function, in our Ceres-based implementation we found that using a Euclidean error term improved the problem's convergence properties.
This is likely because the distribution of cosine distances is not well described by a zero-mean Gaussian distribution (see \Cref{fig:az-zen-err}).

In principle, \Cref{eq:cost_function,eq:cost_function_with_sun} could include an additional term to account for uncertainty in the transformation $\mbf{T}_{b,w}$, which was previously an estimated quantity.
Although the omission of this term means that our estimator may be under-confident in the sun measurements for certain segments of the trajectory, we found that a well chosen static covariance on the sun measurements nevertheless produced good results in practice.
We therefore defer an investigation of this more principled uncertainty propagation to future work.

\subsection{Visual Illumination Estimation} \label{sec:illumination_estimation}
While Lambert et al.~\cite{Lambert2012-um} make use of a hardware sun sensor to estimate the direction of the sun relative to the vehicle, in our approach we wish to use the existing RGB image stream to compute this illumination information in addition to the motion of the camera.
We examine three techniques for estimating the sun direction in a single outdoor RGB image: the technique of Lalonde et al.~\cite{Lalonde2011-jw}, which estimates the sun direction based on a combination of weak visual cues; an improved version of~\cite{Lalonde2011-jw} that makes use of a novel VO-informed prior term to improve its accuracy and reliability; and Sun-CNN, a recent technique for estimating the sun direction using a Convolutional Neural Network (CNN)~\cite{Ma2016-zl}.

\subsubsection{``Lalonde''~\cite{Lalonde2011-jw}} 
\begin{figure}[t]
    \centering   
    \begin{subfigure}[b]{0.45\textwidth}
        \includegraphics[width=\textwidth]{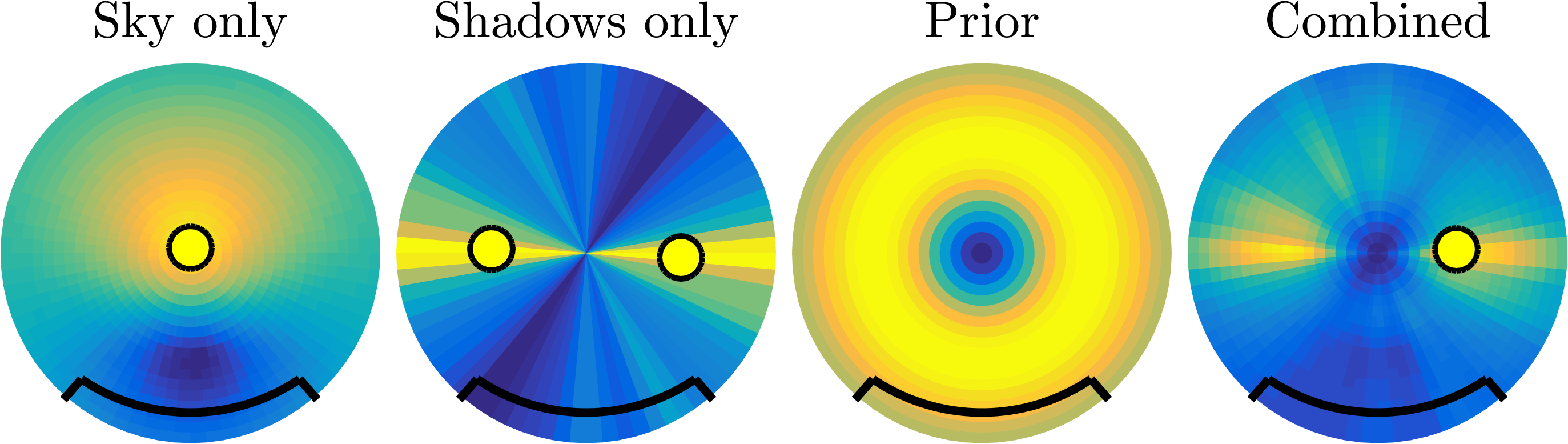}
    \end{subfigure}
    \qquad
    \begin{subfigure}[b]{0.45\textwidth}
        \includegraphics[width=\textwidth]{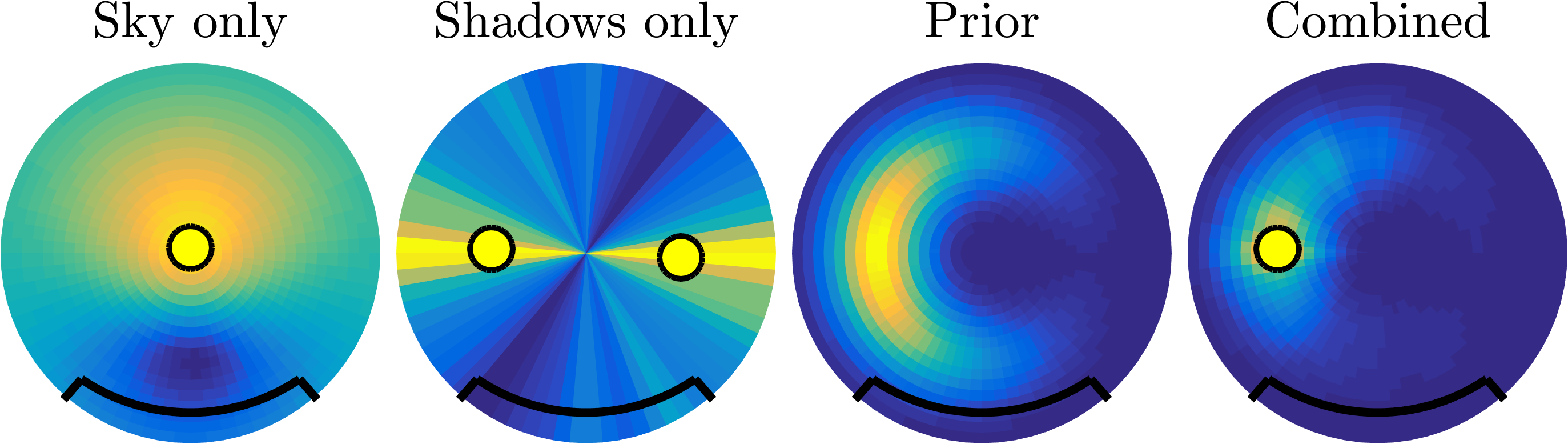}
    \end{subfigure}
    
    \begin{subfigure}[b]{0.45\textwidth}
        \includegraphics[width=\textwidth]{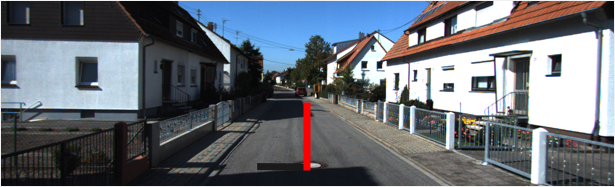}
        \caption{An ambiguous detection resulting in an incorrect maximum likelihood solution, using the prior term of~\cite{Lalonde2011-jw}.}
        \label{fig:2011_09_30_drive_0018_frame_oldprior}
    \end{subfigure}
    \qquad
    \begin{subfigure}[b]{0.45\textwidth}
        \includegraphics[width=\textwidth]{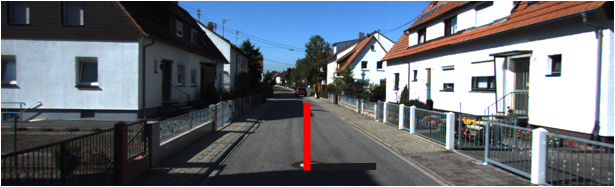}
        \caption{The ambiguity is resolved using a VO-informed prior, which constrains the distribution over sun positions.}
        \label{fig:2011_09_30_drive_0018_frame_newprior}
    \end{subfigure}
    
    \caption{Sample frame from KITTI sequence \texttt{2011\_09\_30\_drive\_0018} and associated sun detection results using the ``Lalonde''~\cite{Lalonde2011-jw} and ``Lalonde-VO'' methods. \emph{Top row}: Probability distributions over sun positions are shown for each visual cue independently, and for the combined result. The maximum likelihood solution(s) are represented as yellow circles, and the camera's field of view is shown in black. \emph{Bottom row}: A virtual sundial (red line) is inserted in the image and casts a virtual shadow (black line) using the detected sun position.}
    \label{fig:2011_09_30_drive_0018_frame}
\end{figure}

estimates the maximum likelihood azimuth-zenith sun direction in a single RGB image by combining relatively weak information from a physically based sky model~\cite{Perez1993-cy}, shadow detection, pedestrian detection, and vertical surface detection routines, as well as a data-driven prior term that captures the distribution of typical sun zeniths in photographs.
An implementation of this technique is freely available as open-source software.\footnote{\url{https://github.com/jflalonde/illuminationSingleImage}}
For our purposes, we use only a subset of these visual cues since the others tended to produce erroneous or null results in our experiments.
Specifically, we use the sky model, shadow detection, and prior term described in~\cite{Lalonde2011-jw}. 
\Cref{fig:2011_09_30_drive_0018_frame_oldprior} shows an example of the results we obtained using this method.
Note that in this case the algorithm produced an incorrect sun detection due to the bimodal ambiguity in the shadow cue and the symmetry of the sky model and prior term.

\begin{figure}[t]
    \centering
    \begin{subfigure}[b]{0.85\textwidth}
        \includegraphics[width=\textwidth]{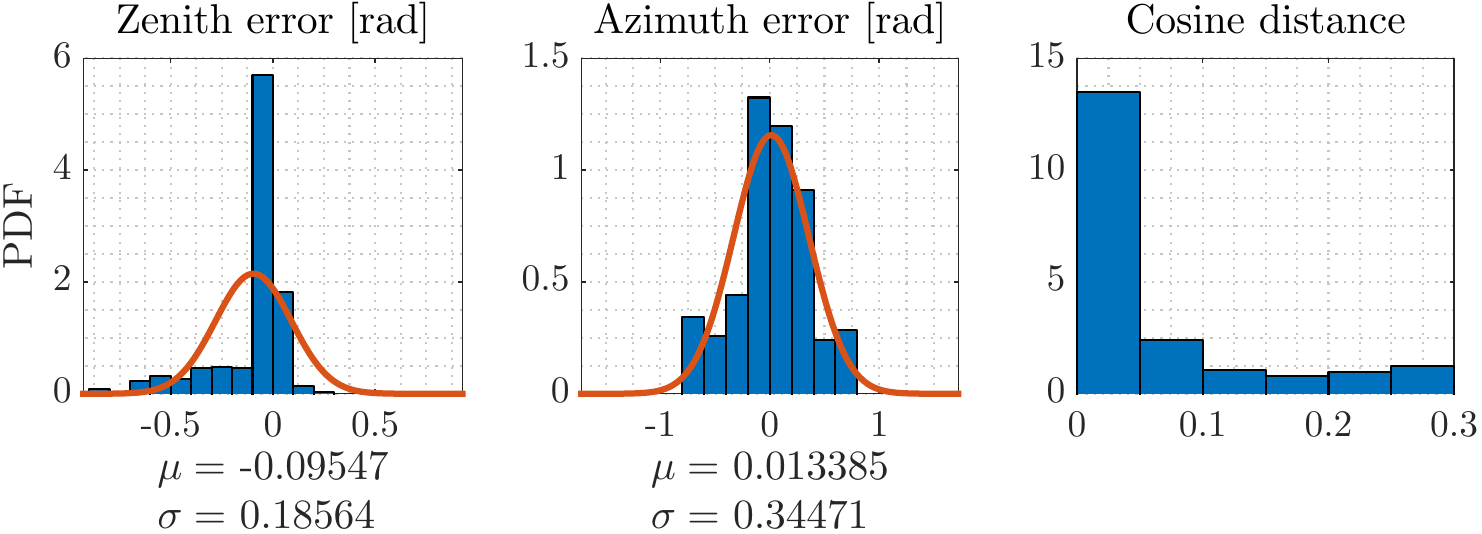}
        \caption{Azimuth errors are approximately zero mean and Gaussian, but zenith errors are skewed due to failures in the sky cue and the biased nature of the prior term in~\cite{Lalonde2011-jw} (\Cref{fig:2011_09_30_drive_0018_frame_oldprior}), which fails to correctly capture the distribution of sun positions in the KITTI dataset.}
        \label{fig:az-zen-err-skew}
        \vspace{10pt}
    \end{subfigure}
    \begin{subfigure}[b]{0.8\textwidth}
        \includegraphics[width=\textwidth]{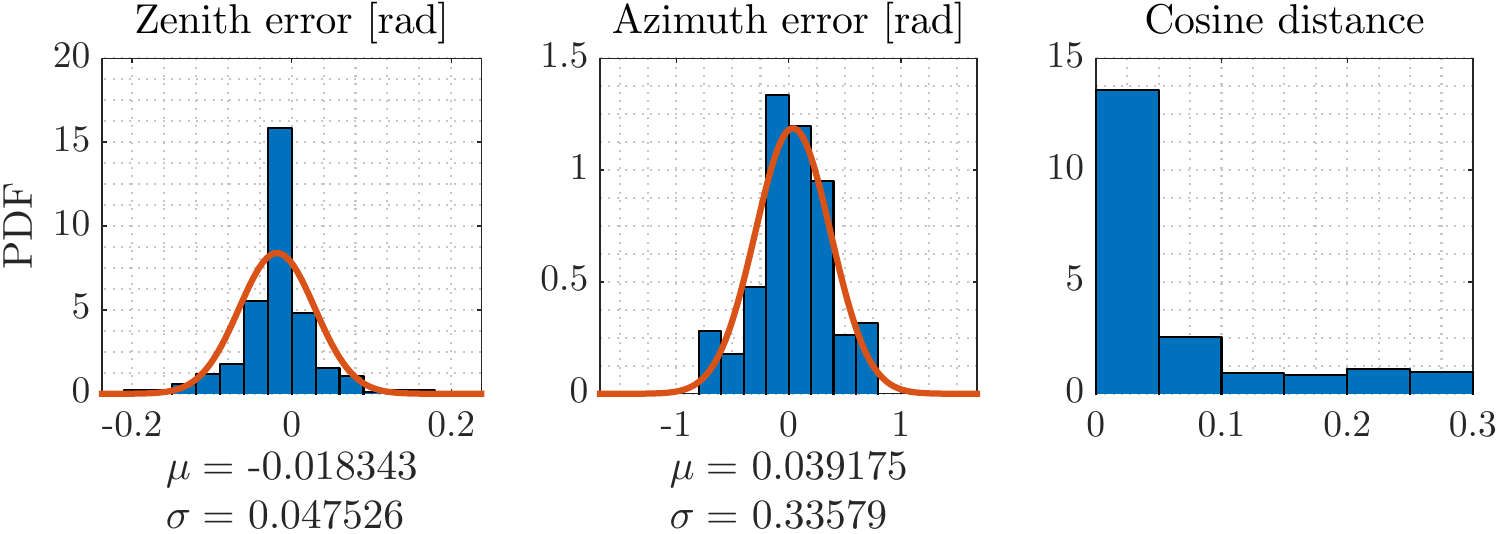}
        \caption{Thresholding the zenith error to exclude the skewed portion of the distribution yields a more Gaussian-like distribution of zenith errors.}
        \label{fig:az-zen-err-unskew}
    \end{subfigure}
    \caption{Distribution of estimation errors for~\cite{Lalonde2011-jw} relative to the ground truth sun vector transformed through the chain of ground truth camera poses. We use a cosine distance threshold of 0.3 to reject outlier estimates.}
    \label{fig:az-zen-err}   
    \vspace{-10pt}
\end{figure}

Since \cite{Lalonde2011-jw} tends to fail in the presence of ambiguous shadows and saturated sky pixels, we reject obvious outliers in our VO pipeline by thresholding the cosine distance between the observed and predicted sun directions based on the current pose estimate.
In practice, we found a cosine distance threshold of 0.3 to be a reasonable choice.
However, as shown in \Cref{fig:az-zen-err-skew}, the distribution of zenith errors is skewed.
This is due to the bias introduced by the prior term of~\cite{Lalonde2011-jw}, which fails to correctly capture the distribution of sun zeniths in the KITTI dataset.
We resolve this issue by thresholding the zenith error (or, equivalently, the $y$-component error in the camera frame) to exclude the skewed portion of the distribution, yielding a more Gaussian-like distribution over zenith errors.

\subsubsection{``Lalonde-VO''} is a modified version of~\cite{Lalonde2011-jw} where we have replaced the original zenith-only prior term with a novel prior term that incorporates the expected sun direction based on the current VO estimate.
The motivation for incorporating this information is twofold.
First, in cases where the sky cue fails, the shadow cue's bimodal probability distribution forces the algorithm to choose one of the two possible solutions at random, leading to a high proportion of erroneous measurements (\Cref{fig:2011_09_30_drive_0018_frame_oldprior}).
By incorporating a weak prior based on the estimated camera pose, we can resolve the ambiguity in the two solutions (\Cref{fig:2011_09_30_drive_0018_frame_newprior}).
Second, ambiguous shadow cues often result in an incorrect pair of maximum likelihood sun azimuths, yet there is typically a secondary pair of local maxima with lower probability that are in fact correct. 
The sky cue alone is not generally strong enough to bias the result towards the correct direction in these cases, but our new VO-informed prior term allows the algorithm to ignore incorrect shadow orientations and incorporate information from the weaker pair of maxima.

We define our VO-informed prior term as a Gaussian distribution over azimuth and zenith angles whose mean is the expected sun direction, and choose the covariance of this distribution such that the $3\sigma$ bounds on the azimuth prior span $360^\circ$, while the $3\sigma$ bounds on the zenith prior span $90^\circ$.
In this way, we account for uncertainty in the camera poses and avoid excessively biasing the sun detection; we need only bias the result towards the correct `half' of the sky.

\subsubsection{``Sun-CNN''~\cite{Ma2016-zl}} uses a Convolutional Neural Network (CNN) trained on sequences from the KITTI dataset \cite{Geiger2013-ky} annotated with ground truth sun directions to estimate the likely azimuth angle of the sun from a single RGB image.
Ma et al.~\cite{Ma2016-zl} show that Sun-CNN substantially outperforms~\cite{Lalonde2011-jw} in terms of azimuth estimation accuracy on the KITTI odometry benchmark, but since it does not estimate the zenith angle of the sun, it is best suited to planar navigation tasks such as autonomous driving of land vehicles.
Since our sun-corrected VO pipeline requires the full 3D direction of the sun relative to the camera, we assign a value of zero and a large covariance to the vertical component of the Sun-CNN estimate so that the unknown component of the sun direction is effectively ignored.

\begin{figure}[b]
    \centering
    \includegraphics[width=\textwidth]{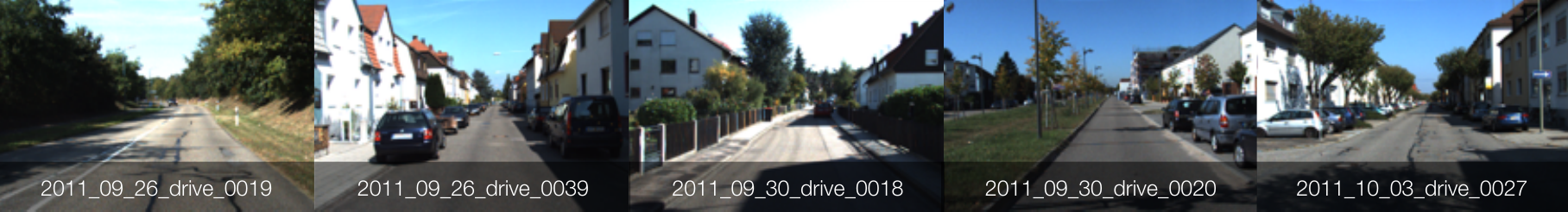}
    \caption{Sample frames from five sequences from the KITTI raw dataset~\cite{Geiger2013-ky}, ranging in length from 300 m to 3.7 km. These sequences contain strong shadows and mostly unsaturated skies, which are amenable to visual sun direction estimation.}    
    \label{fig:kitti_frames}
\end{figure}

\section{Results} \label{sec:results}
We present results for a combined 7.8 km of urban driving from the popular KITTI dataset~\cite{Geiger2013-ky} using a two-frame sliding window and estimated sun directions from each algorithm for every fifth image.
\Cref{fig:kitti_frames} shows sample frames from five sequences in the KITTI raw dataset, ranging in length from 300 m to 3.7 km, which we selected mainly for their strong shadows and unsaturated sky pixels.
We evaluate the translational and rotational average root mean squared error (ARMSE) and the final translational drift error of our VO algorithm, both with and without the sun-based orientation correction.

We processed each sequence using the same set of stereo feature tracks obtained from \texttt{libviso2}~\cite{Geiger2011-xe}, first using pure VO alone, then by incorporating measurements from each sun detection method in turn.
The covariances associated with each sun detection algorithm were individually tuned to reflect the measurement error distribution of each algorithm, and we made a bona fide effort to present the best performance of each algorithm on each sequence.

\begin{figure}
    \centering
    \begin{tikzpicture}
        \node[anchor=south west,inner sep=0] at (0,0) {\includegraphics[width=0.7\textwidth]{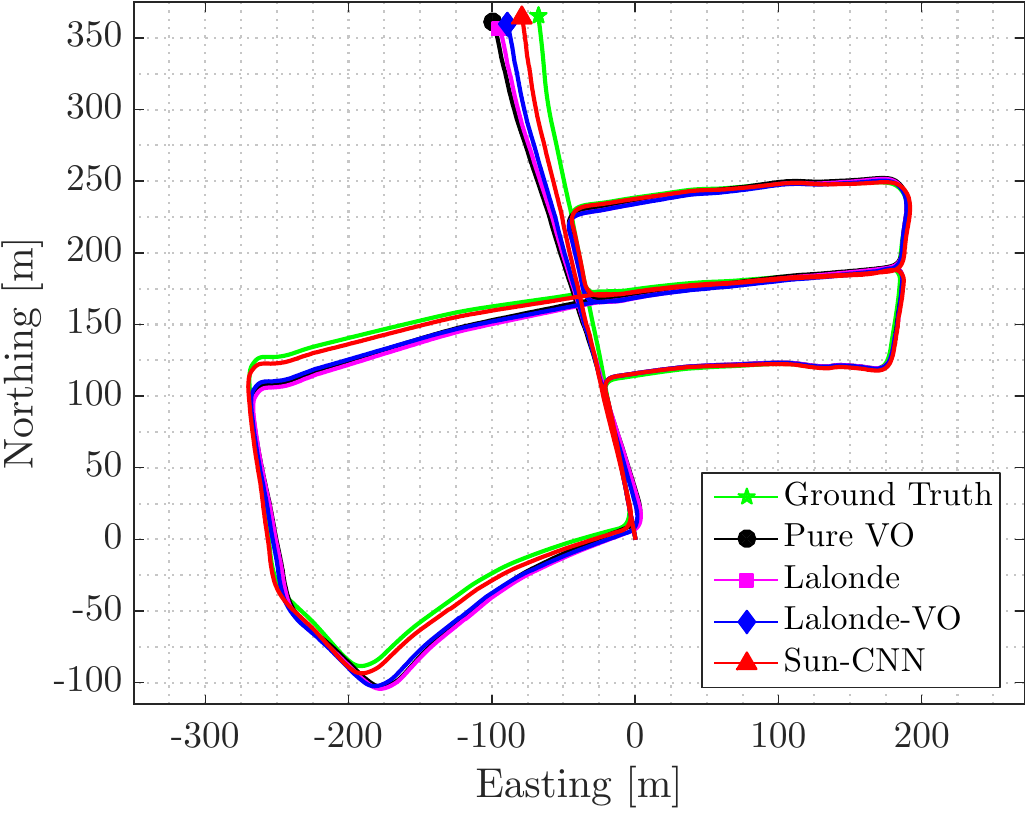}};
        \draw[-latex, black, thick] (4.3,2.7) -- (5.3,2.3);
        \node at (3.9,2.9) {Start};
        \draw[-latex, black, thick] (5.5,6.2) -- (4.5,6.6);
        \node at (5.9,6.1) {End};
    \end{tikzpicture}
    \caption{VO results for the 2.2 km sequence \texttt{2011\_09\_30\_drive\_0018}. The VO result is visibly closer to ground truth using the sun-based orientation correction.}
    \label{fig:2011_09_30_drive_0018_traj}
    \vspace{-10pt}
\end{figure}
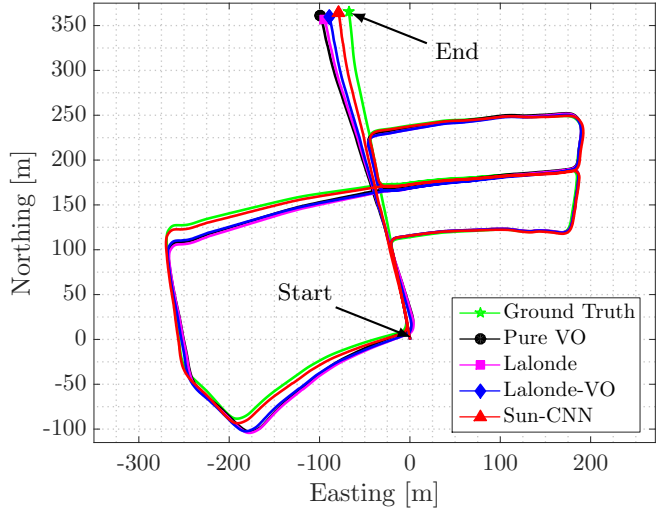

\Cref{fig:2011_09_30_drive_0018_traj} shows the estimated and ground truth trajectories for the 2.2 km sequence \texttt{2011\_09\_30\_drive\_0018}.
With the exception of the ``Lalonde'' method, the sun-aided VO trajectories are noticeably closer to ground truth than the pure VO trajectory.
The ``Lalonde'' method appears to have had minimal impact on this sequence due to the relatively low number of inlier sun detections.

\begin{table}
\centering
\caption{Average Root Mean Squared Error (ARMSE) and Final Translational Drift Error on KITTI Sequences}
\begin{tabular}{lccccc}
& && \multicolumn{3}{c}{VO + Sun Estimation} \B \\ \cline{4-6}
& Pure VO && Lalonde & Lalonde-VO & Sun-CNN \T\B \\ \hline
\texttt{2011\_09\_26\_drive\_0019} (0.4 km) & \T\B \\
\quad Trans. ARMSE [m] & 4.99 && \textbf{4.93} & 4.94 & 5.20 \T \\
\quad Trans. ARMSE (EN-plane) [m] & \textbf{5.42} && 5.52 & 5.52 & 5.49 \\
\quad Rot. ARMSE ($\times 10^{-3}$) [axis-angle] & \textbf{1.47} && 1.61 & 1.61 & 1.90 \B \\
\quad Final trans. drift [m] & 13.04 && \textbf{12.83} & 12.89 & 13.88 \T \\
\quad Final trans. drift [\%] & 3.21 && \textbf{3.16} & 3.18 & 3.42 \\
\quad Final trans. drift (EN-plane) [m] & \textbf{11.45} && 11.74 & 11.77 & 11.74 \\
\quad Final trans. drift (EN-plane) [\%] & \textbf{2.82} && 2.89 & 2.90 & 2.89 \B \\
\texttt{2011\_09\_26\_drive\_0039} (0.3 km) & \T\B \\
\quad Trans. ARMSE [m] & 2.51 && 2.50 & \textbf{2.48} & 2.53 \T \\
\quad Trans. ARMSE (in-plane) [m] & \textbf{2.53} && 2.55 & 2.54 & 2.57 \\
\quad Rot. ARMSE ($\times 10^{-3}$) [axis-angle] & 1.08 && 1.13 & 1.14 & \textbf{0.06} \B \\
\quad Final trans. drift [m] & 8.14 && 8.01 & \textbf{7.98} & 8.40 \T \\
\quad Final trans. drift [\%] & 2.74 && 2.69 & \textbf{2.68} & 2.82 \\
\quad Final trans. drift (EN-plane) [m] & 6.69 && 6.77 & \textbf{6.65} & 7.01 \\
\quad Final trans. drift (EN-plane) [\%] & 2.26 && 2.27 & \textbf{2.23} & 2.35 \B \\
\texttt{2011\_09\_30\_drive\_0018} (2.2 km) & \T\B \\
\quad Trans. ARMSE [m] & 4.66 && 6.68 & 5.47 & \textbf{2.67} \T \\
\quad Trans. ARMSE (EN-plane)[m] & 5.43 && 5.95 & 5.00 & \textbf{2.09} \\
\quad Rot. ARMSE ($\times 10^{-3}$) [axis-angle] & 3.52 && 5.71 & 4.65 & \textbf{2.23} \B \\
\quad Final trans. drift [m] & 32.67 && 31.74 & 26.35 & \textbf{13.44} \T \\
\quad Final trans. drift [\%] & 1.48 && 1.44 & 1.20 & \textbf{0.61} \\
\quad Final trans. drift (EN-plane) [m] & 31.45 && 28.00 & 22.18 & \textbf{11.33} \\
\quad Final trans. drift (EN-plane) [\%] & 1.43 && 1.27 & 1.01 & \textbf{0.51} \B \\
\texttt{2011\_09\_30\_drive\_0020} (1.2 km) & \T\B \\
\quad Trans. ARMSE [m] & 3.07 && 3.21 & 3.03 & \textbf{2.94} \T \\
\quad Trans. ARMSE (EN-plane) [m] & 3.37 && 3.51 & 3.35 & \textbf{3.34} \\
\quad Rot. ARMSE ($\times 10^{-3}$) [axis-angle] & 2.10 && 2.42 & 2.64 & \textbf{1.69} \B \\
\quad Final trans. drift [m] & 7.19 && 7.47 & \textbf{6.57} & 7.23 \T \\
\quad Final trans. drift [\%] & 0.58 && 0.61 & \textbf{0.53} & 0.59 \\
\quad Final trans. drift (EN-plane) [m] & 6.43 && \textbf{6.00} & 6.52 & 7.23 \\
\quad Final trans. drift (EN-plane) [\%] & 0.52 && \textbf{0.49} & 0.53 & 0.58 \B \\
\texttt{2011\_10\_03\_drive\_0027} (3.7 km) & \T\B \\
\quad Trans. ARMSE [m] & 4.10 && 13.84 & 10.63 & \textbf{4.08} \T \\
\quad Trans. ARMSE (in-plane) [m] & 4.20 && 3.53 & \textbf{2.57} & 4.27 \\
\quad Rot. ARMSE ($\times 10^{-3}$) [axis-angle] & 2.28 && 9.31 & 4.97 & \textbf{2.20} \B \\
\quad Final trans. drift [m] & 10.06 && 13.35 & \textbf{8.31} & 8.96 \T \\
\quad Final trans. drift [\%] & 0.27 && 0.36 & \textbf{0.22} & 0.24 \\
\quad Final trans. drift (EN-plane) [m] & 8.33 && \textbf{2.53} & 4.23 & 8.30 \\
\quad Final trans. drift (EN-plane) [\%] & 0.22 && \textbf{0.07} & 0.11 & 0.22 \B \\ \hline
\end{tabular}  
\label{tab:armse}
\end{table}

\Cref{tab:armse} quantifies the difference in the each result by reporting their translational and rotational ARMSE, as well as the final translational drift error, relative to ground truth.
We see that including the sun-based orientation correction can yield a substantial reduction in estimation error compared to pure VO, particularly on the longer sequences, which contain several sharp turns.
This is especially apparent in the case of sequence \texttt{2011\_09\_30\_drive\_0018}, which enjoys a 43\% reduction in translational ARMSE (62\% in-plane), and a 59\% reduction in final translational drift error (64\% in-plane) using the Sun-CNN method~\cite{Ma2016-zl}.
We stress that this improvement is purely due to information already available in the existing image stream -- no additional sensors are required.
On the other hand, short straight sequences such as \texttt{2011\_09\_26\_drive\_0019} and \texttt{2011\_09\_26\_drive\_0039} do not benefit significantly from sun measurements since the accumulated orientation error in the VO estimate is already small.

Overall, the ``Sun-CNN'' and ``Lalonde-VO'' methods outperform the ``Lalonde'' method in terms of reducing estimation error in our stereo VO pipeline.
This is to be expected since the ``Lalonde-VO'' method incorporates additional information about the temporal consistency of the images, while Ma et al.~\cite{Ma2016-zl} have already shown that Sun-CNN is both more accurate and more reliable than~\cite{Lalonde2011-jw} on single images in the KITTI dataset.
While ``Sun-CNN'' and ``Lalonde-VO'' yield the minimum estimation error in similar numbers of cases, in the cases where ``Sun-CNN'' performs better, it does so by a wide margin.
Furthermore, Sun-CNN is faster to evaluate than the other two algorithms while simultaneously avoiding hand-crafted features and approximate models of hand-picked cues. 
This suggests that high level scene understanding using machine learning may be a promising tool for improving robot localization accuracy in addition to providing semantic information about the environment.

\section{Conclusions and Main Experimental Insights}
In this work we have shown that estimation error in stereo visual odometry (VO) can be reduced by exploiting global illumination information available from the same image stream used to compute VO.
The main insight here is that there is much to be gained in visual navigation by reasoning over more than just geometry.
In particular, the notion of embracing illumination as a tool for localization is one that has not been widely adopted, yet is a promising direction for future research.
Convolutional Neural Networks (CNNs) in particular appear to be excellent tools for extracting illumination information in a form amenable to conventional VO techniques.
Future work might focus on developing these tools further to yield even greater gains in localization accuracy and robustness.

\bibliographystyle{abbrvurl}
\bibliography{bib,ceres}

\end{document}